\documentclass[a4paper,12pt,singlecolumn]{elsarticle}
\usepackage[top=3cm, bottom=3cm, left=3cm, right=3cm]{geometry}
\usepackage{setspace}
\usepackage{mathptmx}
\usepackage{amsmath,amssymb,amsfonts}
\usepackage{algorithmic}
\usepackage{graphicx}
\usepackage{textcomp}
\usepackage{xcolor}
\usepackage{url}
\usepackage{algorithmic}
\usepackage[ruled]{algorithm2e}
\usepackage{tikz,pgfplots}
\usepackage{pgfplotstable}
\usepackage{subfig}
\usepackage{pifont}
\usepackage{multirow,booktabs}
\usepackage{booktabs}
\usepackage{caption}

\usepackage{float}
\restylefloat{figure}
\restylefloat{table}

\begin{document}

\begin{frontmatter}
		
		\title{AnomalyAID: Reliable Interpretation for Semi-supervised Network Anomaly Detection}



\tnotetext[2]{Corresponding author(s): Yingwen Wu and Jin Wang}

\author[1]{Yachao Yuan}
\ead{chao910904@suda.edu.cn}
\affiliation[1]{organization={School of Future Science and Engineering, Soochow University},
            city={Suzhou},
            state={Jiangsu},
            country={China}}


\author[3]{Yu Huang}
\ead{fatmo@nuaa.edu.cn}
\affiliation[3]{organization={Southeast University},
            city={Nanjing},
            state={Jiangsu},
            country={China}}

\author[1]{Yingwen Wu*}
\ead{ywwu@suda.edu.cn}

\author[1]{Jin Wang*}
\ead{ustc\_wangjin@hotmail.com}

\begin{abstract}
Semi-supervised Learning plays a crucial role in network anomaly detection applications, however, learning anomaly patterns with limited labeled samples is not easy. Additionally, the lack of interpretability creates key barriers to the adoption of semi-supervised frameworks in practice. Most existing interpretation methods are developed for supervised/unsupervised frameworks or non-security domains and fail to provide reliable interpretations. In this paper, we propose AnomalyAID, a general framework aiming to (1) make the anomaly detection process interpretable and improve the reliability of interpretation results, and (2) assign high-confidence pseudo labels to unlabeled samples for improving the performance of anomaly detection systems with limited supervised data. For (1), we propose a novel interpretation approach that leverages global and local interpreters to provide reliable explanations, while for (2), we design a new two-stage semi-supervised learning framework for network anomaly detection by aligning both stages' model predictions with special constraints. We apply AnomalyAID over two representative network anomaly detection tasks and extensively evaluate AnomalyAID with representative prior works. Experimental results demonstrate that AnomalyAID can provide accurate detection results with reliable interpretations for semi-supervised network anomaly detection systems. 
\end{abstract}




\end{frontmatter}


\section{Introduction}
Anomaly detection has become a fundamental task in various applications, including network intrusion detection \cite{mirsky2018kitsune}, web attack detection \cite{9155278}, and advanced persistent threat detection \cite{259729}. Anomaly detection systems aim to identify unforeseen threats, such as zero-day attacks. To achieve this, semi-supervised learning is increasingly promising, as it requires only limited labeled data, unlike traditional supervised approaches that depend on extensive labeled samples for training.

Despite the great promise of semi-supervised network anomaly detection systems, the lack of interpretability of their predictions poses major barriers to their practical adoption. Firstly, it is challenging to build trust in the decisions made by these systems, which often provide only binary outputs (benign or malicious) without adequate justification or reliable evidence. Second, it is nearly impossible for analysts to manually interpret the decisions of these black-box machine learning models, as the volume of training data and the complexity of the learned models are beyond human understanding \cite{guidotti2018survey}. Third, minimizing false positives and miss-detected samples remains a critical challenge for anomaly detection systems in real-world applications. Without a clear understanding of how the models operate, it is impossible to effectively update or adjust them to reduce them \cite{han2021deepaid}. Consequently, security operators are left unable to decide whether they can trust model decisions based on excessively simple model predictions, are unwilling to use theoretical semi-supervised network anomaly detection systems in their practical applications, and feel overwhelmed by numerous meaningless false positives.

\begin{figure}[t!]
    \centering
    \includegraphics[width= 0.9\linewidth]{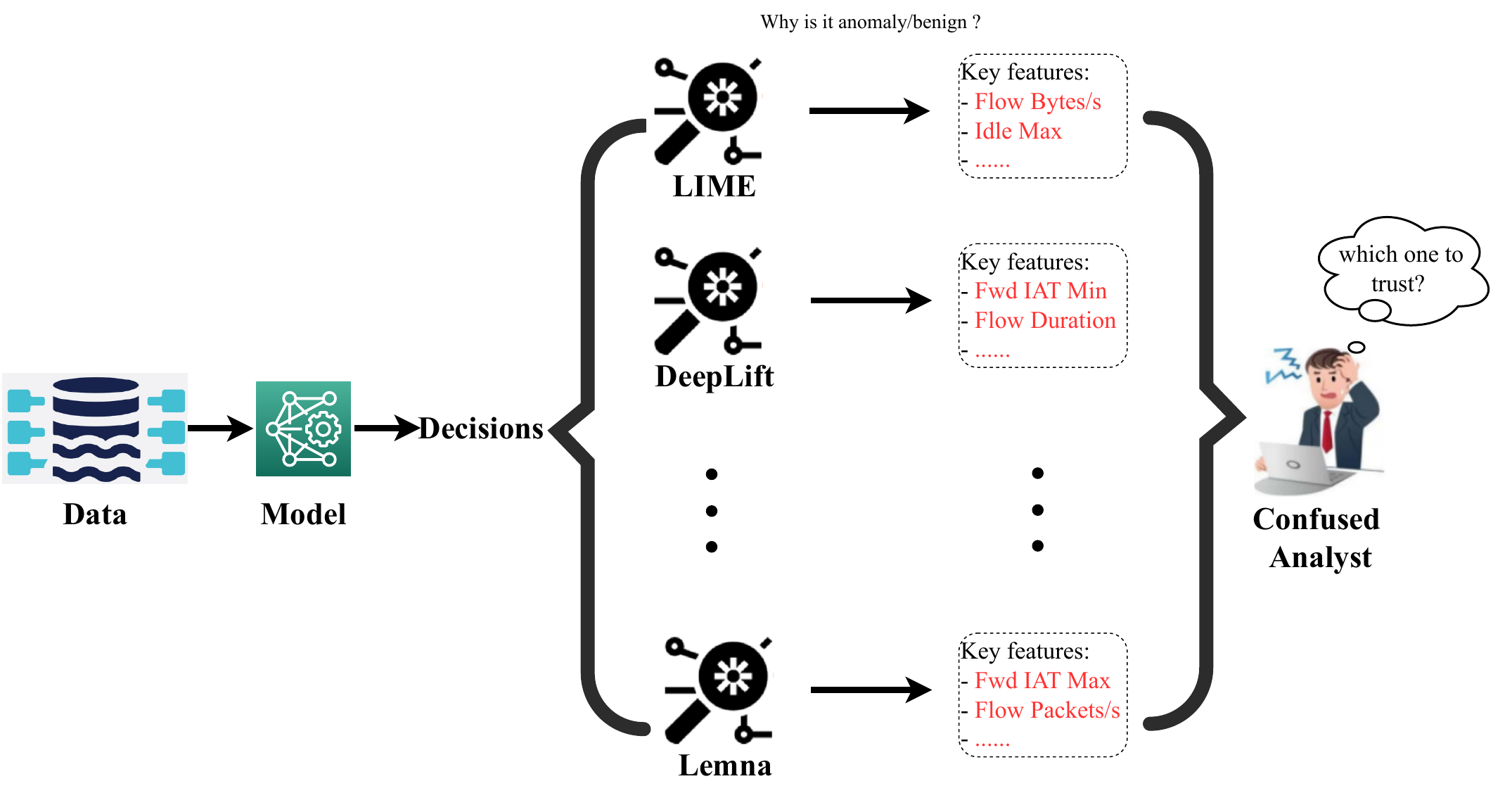}
    \caption{Illustration of the divergent explanation results of different interpreters.}
    \label{fig:problem}
    \vspace{-5mm}
\end{figure}

Recently, some research has proposed techniques for interpreting machine learning model decisions by identifying key features that significantly influence the final prediction \cite{ribeiro2016should_lime,lundberg2017unified_shap,guo2018_lemna}. Predominant approaches include local/model-agnostic \cite{ribeiro2016should,ribeiro2018anchors,lundberg2017unified_shap,guo2018_lemna} and global \cite{fisher2019all,9817459} methods that focus on interpreting individual and cluster predictions for a given black-box classifier. These interpreters either try to approximate the local decision boundary using a linear algorithm to find key features of the given input, or they perturb the current input's feature values or permute features of a cluster and observe the model’s feedback to pinpoint the most influential ones. Security operators can use these interpreted key features to better understand model decisions \cite{guidotti2018survey}. 
Existing work like \cite{fong2017interpretable,shrikumar2017learning,9817459,mothilal2020explaining,shrikumar2017learning_deeplift,han2021deepaid,wei2023xnids} predominantly focuses on interpreting supervised/unsupervised models in various applications. However, the field of developing reliable interpretation for semi-supervised network anomaly detection remains unexplored.
Fig. \ref{fig:problem} illustrates the process of a classifier analyzing network traffic data and producing decisions, such as identifying whether the traffic is anomalous or benign. Various interpretability methods, including LIME \cite{ribeiro2016should_lime}, DeepLIFT \cite{shrikumar2017learning_deeplift}, and Lemna \cite{guo2018_lemna}, generate different key features (e.g., "Flow Bytes/s," "Idle Max," and "Fwd IAT Min"), leading to confusion for the security analyst. The inconsistency in these explanations raises uncertainty regarding the reliability of the interpretations.

\textbf{Our Work.} In this paper, we propose AnomalyAID, a general semi-supervised framework for automatically learning from enormous unlabeled data and improving the reliability of interpretation results. The design goal of AnomalyAID is to develop a novel semi-supervised learning framework for network anomaly detection applications that meets the special requirements of security domains  (such as reliability). To this end, we propose two techniques in AnomalyAID referred to as Global-local Knowledge Association Mechanism (KAM) and Two-stage Semi-supervised Learning System (ToS). KAM helps security analysts understand the underlying reason for model predictions, while ToS provides an effective pseudo-labeling process to assist the semi-supervised learning. 
Compared to original black-box machine learning models, security practitioners can better understand system feedback. Besides, with more reliable interpretations, system operators are more willing to adopt theoretical semi-supervised frameworks in their practical applications.
We also provide prototype implementations of AnomalyAID over two representative network anomaly detection datasets (ISCXTor2016 \cite{lashkari2017characterization}, CIC-DoHBrw-2020 \cite{montazerishatoori2020detection}, and UNSW-NB15 \cite{moustafa2015unsw}). The proposed KAM and ToS are extensively evaluated with representative prior approaches. Experimental results demonstrate that AnomalyAID can provide reliable explanations for semi-supervised learning frameworks while satisfying the requirements of security-related applications (Our code is available at: https://github.com/M-Code-Space/AnomalyAID.).

\textbf{Contributions.} This paper has the following contributions:
\begin{itemize}
    \item We propose AnomalyAID, a general framework for interpreting and improving semi-supervised network anomaly detection in security-related applications, which includes two key components: KAM provides reliable interpretations for semi-supervised systems and ToS makes it possible to learn from large unlabeled datasets automatically.
    \item We conduct extensive experiments to demonstrate that the proposed AnomalyAID outperforms existing approaches regarding fidelity, stability, robustness, and efficiency. 
\end{itemize}

\section{Motivating Scenarios}
\begin{figure}[t!]
    \centering
    \includegraphics[width=0.8\linewidth]{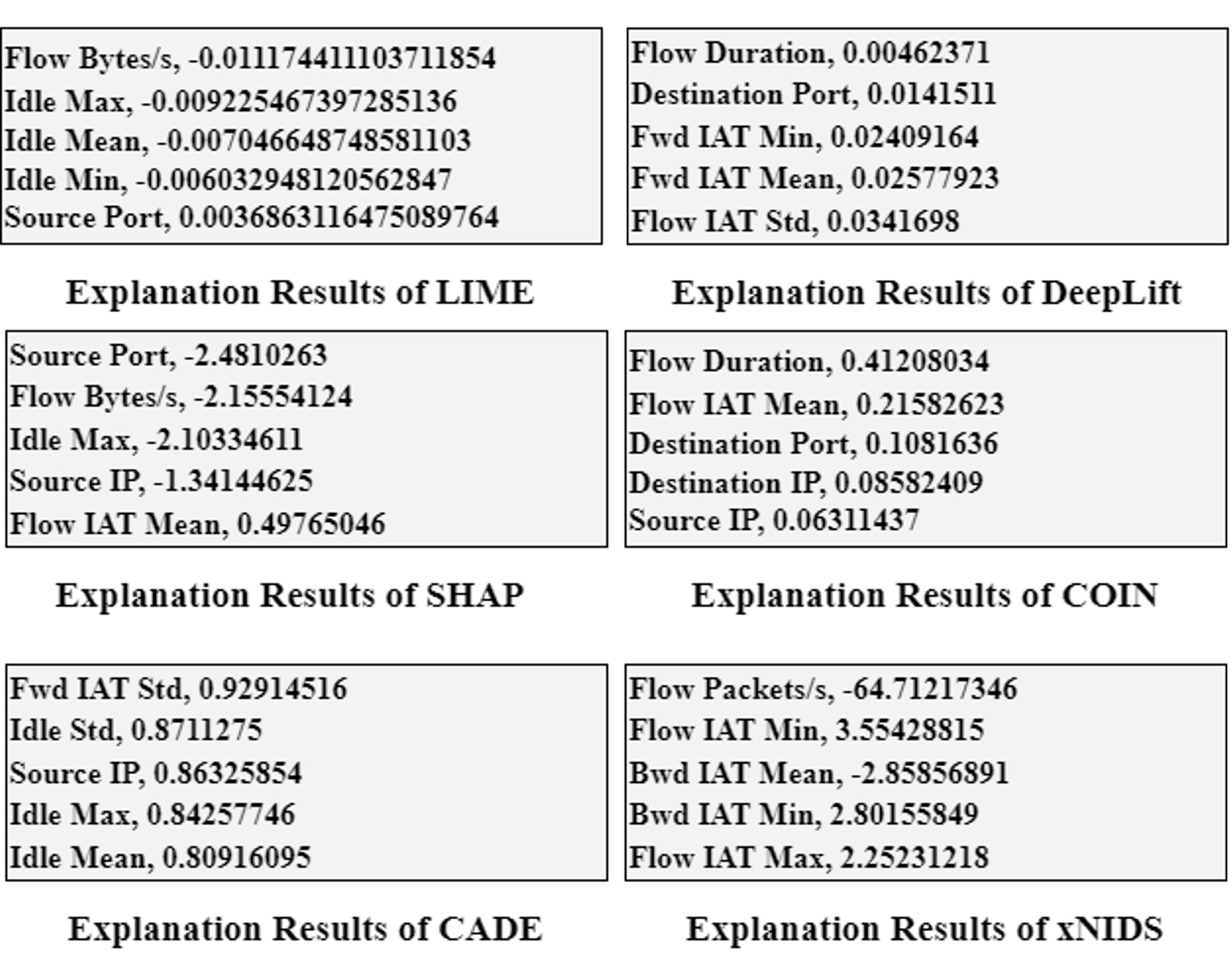}
    \caption{Six example explanation results for the same input data with the same
 model using LIME, DeepLift, SHAPE \cite{lundberg2017unified_shap}, COIN \cite{liu2017contextual_coin}, CADE \cite{yang2021cade}, and xNIDS \cite{wei2023xnids}. Different interpreters give different explanations.}
    \label{fig:explanations}
\end{figure}

For network anomaly detection applications, a security analyst relies on a machine learning-based algorithm to identify network anomalies. The machine learning model is trained in a supervised or semi-/un-supervised manner. After training the detector, the security analyst analyzes incoming samples and classifies them as either malicious or benign. However, since the model operates as a black box, it doesn’t provide any reasoning behind its decisions. To understand why the system flagged a sample, the analyst turns to explainability methods that highlight critical features of the input. Fig. \ref{fig:explanations} shows the interpretation outcomes for an input sample identified as anomalies by the same detector, employing six distinct local interpreters: LIME, DeepLift, SHAPE, COIN, CADE, and xNIDS. 


In Fig. \ref{fig:explanations}, the interpretation results of LIME \cite{ribeiro2016should_lime}, DeepLift \cite{shrikumar2017learning_deeplift}, SHAPE \cite{lundberg2017unified_shap}, COIN \cite{liu2017contextual_coin}, CADE \cite{yang2021cade}, and xNIDS \cite{wei2023xnids} highlight five key features (when the security analyst sets the number of key features as five). Note that the numerical values associated with these features represent their respective impact weights. Interestingly, the outputs from the six methods exhibit significant differences. For instance, LIME and DeepLift show no overlapping features in their results. Additionally, while the Source IP feature in SHAP has a negative influence on the classification of an input network traffic instance, in COIN, it demonstrates an opposing effect. A similar phenomenon can be observed for the Source Port feature. These discrepancies leave the analyst perplexed. He/she may wonder which method to select and whether these explanations can be relied upon. Given the lack of standardized metrics for assessing explainability techniques in network anomaly detection, this situation underscores the need for an evaluation study to assess the reliability and applicability of these methods in this critical area.



\textbf{Network anomaly detection.}
In the process of detecting network anomalies (i.e., classifying network traffic into malicious or benign traffic), each network traffic flow $x_i$ is represented as a feature vector in a $d$-dimensional space, where $x_i \in \mathbb{R}^d$. A network anomaly detector $f$ can be developed using a designated supervised or semi-/un-supervised machine learning algorithm based on this dataset. We can express the prediction for the $i$-th sample as $f(x_i) = \hat{y}_i$, where $\hat{y}_i$ reflects the predicted output associated with $x_i$. The classifier is considered to perform correctly if the predicted label $\hat{y}_i$ aligns with the actual label $y_i$.

\textbf{Interpretation.}
The interpretation process is denoted as $g = m(f)$, where $f$ is a specific pre-trained network anomaly detector and $m$ is an interpreter. Given a network traffic flow $x_i$, the interpreter $m$ will produce the interpretation result $g$, which is represented as a list of features that are ranked based on their contribution to the final decision.

\section{Preliminaries}
\subsection{Explanation methods}
\textbf{PFIE.}
PFIE, proposed by \cite{fisher2019all}, is a simple and easy-to-implement feature importance calculation method used to explain the predictions of machine learning models. Its core idea is to evaluate the importance of a feature to the model's predictions by observing the change in the model's prediction error after shuffling (or permuting) the values of that feature. Specifically, the calculation of PFIE first involves computing the baseline error using the trained model and the original dataset. Then, a specific feature is selected, and its values are randomly shuffled to break the relationship between the feature and the target variable. Next, the model's prediction error is recalculated using the shuffled dataset. Finally, the importance of the feature is assessed by comparing the baseline error with the shuffled error (as defined in Eq., \eqref{eq:PFIE}). The greater the increase in error, the more important the feature is to the model's predictions.
\begin{equation} \label{eq:PFIE}
    \textit{Importance}(f) = \textit{Error}(M) - \textit{Error}(M_p), 
\end{equation}
where $\textit{Importance}(f)$ means the permutation importance of feature $f$, indicating the impact of the feature $f$ on the model's performance. The larger the value, the more important the feature. $\textit{Error}(M)$ is the error metric of model $M$ on the original data, representing the model's performance without modifying the feature. $\textit{Error}(M_p)$ denotes the error metric of the shuffled model $M_p$ on the dataset, reflecting the change in model performance after shuffling feature $f$.

\textbf{DeepLift.} It is introduced by \cite{shrikumar2017learning_deeplift}, which is designed to interpret the output predictions of neural networks by comparing the activation of each neuron to its “reference activation” and decomposing the contributions of all neurons in the network to the input features through backpropagation. In the context of network traffic anomaly detection, the process begins with selecting a zero-input as the reference input to compute the difference in activations. Typically, the reference input is chosen to be a zero vector (i.e., all feature values are set to zero), representing a state of “no features.” For instance, \( x_0 = 0 \). Next, both the reference input \( x_0 \) and the actual input \( x_i \) are propagated through the deep learning-based network traffic anomaly detection model to compute the reference activation \( y_0 \) and the actual activation \( y_i \) for each neuron. The difference between the reference activations and the actual activations is then calculated for each neuron. Subsequently, the rules of DeepLift—such as the Linear Rule, Rescale Rule, or RevealCancel Rule—are applied to assign contribution scores for each input feature (e.g., pixel) to the output. Finally, the contribution scores are backpropagated from the output layer to the input layer using the chain rule. Through this approach, DeepLift provides a clear interpretation of how the CNN model utilizes input features for decision-making, thereby enhancing the model's interpretability. 

\subsection{Machine learning-based classifiers}
\textbf{DBSCAN.} Density-Based Spatial Clustering of Applications with Noise (DBSCAN) \cite{ester1996density} is a popular clustering algorithm that focuses on identifying clusters based on the density of data points. Unlike traditional clustering methods like K-Means, DBSCAN does not require the number of clusters to be specified in advance and can discover clusters of arbitrary shapes while effectively handling noise.

Specifically, it calculates the density around each sample using two parameters: the neighborhood radius $\epsilon$ and the minimum number of points $MinPts$ required to form a dense region. The Euclidean distance between two samples $G(x_i)$ and $G(x_k)$ is denoted as $\|G(x_i) - G(x_k)\|$. The $\epsilon$-neighborhood of a sample $x_i$ is defined as:
\begin{equation}
    N_\epsilon(x_i) = \{x_j \mid \|G(x_i) - G(x_j)\| \leq \epsilon \},
\end{equation}
where $N_\epsilon(x_i)$ represents the set of points within the $\epsilon$-neighborhood of $x_i$. If the number of points in the neighborhood, $|N_\epsilon(x_i)|$, is at least $MinPts$, the sample $x_i$ is classified as a core point. Samples reachable from a core point but not meeting the density requirement are labeled as border points, while samples that are not reachable from any core point are treated as outliers.

\textbf{FCN.} A Fully Connected Network (FCN) \cite{rumelhart1986learning} is a type of neural network architecture where each neuron in one layer is connected to every neuron in the next layer. This architecture is commonly used for tasks such as network traffic anomaly detection. The key ideas and principles of an FCN revolve around the transformation of input data through multiple layers of neurons, with each layer learning increasingly abstract representations of the input.

An FCN consists of multiple layers, including input layers, hidden layers, and an output layer. Each layer is fully connected to the next layer, meaning every neuron in one layer is connected to every neuron in the subsequent layer (the calculation is shown in Eq. \eqref{eq:FC}).
\begin{equation} \label{eq:FC}
    \mathbf{z}^{(l+1)} = \mathbf{W}^{(l)} \mathbf{x}^{(l)} + \mathbf{b}^{(l)}, 
\end{equation}
where $\mathbf{W}^{(l)}$ and $\mathbf{b}^{(l)}$ are the weight matrix and bias vector of layer $l$. $\mathbf{x}^{(l)}$ is the input to layer $l$.
Each neuron in an FCN computes a weighted sum of its inputs and applies a non-linear activation function (e.g., ReLU) to introduce non-linearity into the model. This allows the network to learn complex patterns in the data.
FCNs are trained using backpropagation, where the error between the predicted output and the true label is propagated backward through the network to update the weights. This process minimizes a loss function, usually cross-entropy for classification (see Eq. \eqref{eq:CE}).
\begin{equation} \label{eq:CE}
\mathcal{L} = -\sum_i t_i \log(y_i), 
\end{equation}
where $t_i$ is the true label and $y_i$ is the predicted probability for class $i$.

In our study, we use the PFIE and DeepLift as our global and local interpreters, and  FCN and DBSCAN as our local and global models, respectively, as an example for evaluating the effectiveness of the proposed ToS. Other machine learning models can also be chosen accordingly for specific tasks.

\begin{figure*}[t!]
    \centering
    \begin{center}
    \makebox[\linewidth]{\includegraphics[width=.99\linewidth]{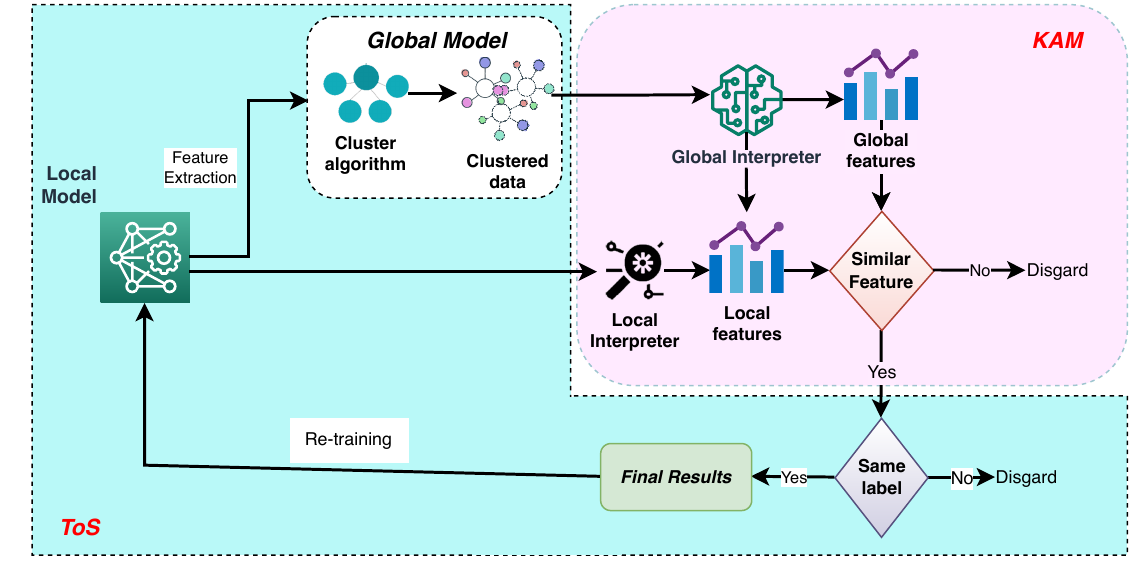}}
    \end{center}
    \caption{Overview of the proposed AnomalyAID.}
    \label{fig:mainFrame}
    \vspace{-4mm}
\end{figure*}

\section{Framework} \label{sec:framework}
\subsection{Overview}
The overview of AnomalyAID is illustrated in Fig. \ref{fig:mainFrame}, consisting of two key components: Global-local Knowledge Association Mechanism (KAM) and Two-stage Semi-supervised Learning System (ToS). ToS enables accurate semi-supervised learning via a novel two-stage pseudo-labeling process, while KAM provides reliable interpretations for the semi-supervised anomaly detection system. AnomalyAID is a general model-independent framework that formulates semi-supervised learning and decision interpretation into a unified end-to-end learning process with security-related constraints. We instantiate AnomalyAID on the aforementioned network anomaly detection security application. With AnomalyAID, malicious network traffic can be automatically detected without requiring large labeled datasets; meanwhile, security practitioners can better understand system behaviors and trust system decisions.

\textbf{Working Process.} 
The working process of the proposed AnomalyAID is as follows. Before starting, we assume that there is a small labeled dataset that can be utilized to pre-train the local and global machine learning models. Such an assumption is reasonable, as semi-supervised learning necessitates the model to continuously assign pseudo-labels to unlabeled samples, and pre-training the model using a small labeled dataset saves data labeling costs while ensuring the effectiveness of the subsequent data-labeling process. Similar assumptions can be found in \cite{li2024instantT}.

Given an instance (or a batch of instances), we fed it into the pre-trained local model to extract its feature embedding and predict its label. The extracted feature embedding is then clustered by the global model into several clusters, where samples from the same cluster share similar characteristics. Feature representations (or feature embeddings, we use them interchangeably in this paper) from a cluster are treated as global representations, while the feature embedding from an instance is defined as a local representation. We group the input instance into one of the clusters by comparing the distance between the embedding of the instance and the clusters. 
Then, the predictions of the global and local models are interpreted by the global and local interpreters, and the reliability of the interpretation results is further improved by comparing the global and local interpretations. 
The underlying reason behind this is that global and local interpreters are heterogeneous interpreters.
It has a higher probability of improving its trustworthiness by matching global and local interpretations instead of two local interpreters.

For instance, in scenarios where individual samples contain noise or have been maliciously altered, the interpretation results at the local level may be wrong; however, due to the fact that the global and local interpreters are heterogeneous, there is a probability that the interpretation results at the global level are correct. In this way, the interpretations' reliability could be improved.
If the interpreted global and local key features match (the similarity score is higher than a given threshold), we consider the key features interpreted by the interpreters to be credible. Conversely, if they do not match, the samples will be discarded. These samples will not be utilized for model re-training to avoid possible model performance deterioration. Finally, labels predicted by the global and local models are aligned to find the most credible pseudo-labels for the next round's semi-supervised training. More details about AnomalyAID are presented in Algorithm~\ref{alg:AnomalyAID}.

\begin{algorithm}[t!]
\caption{AnomalyAID} \label{alg:AnomalyAID}
    \SetKwInOut{Input}{Input}
    \SetKwInOut{Output}{Output}
    \Input{
    Samples $X = \{x_1, \cdots, x_N\}$, local model $M_1$ and global model $M_2$, local interpreter $L$, and global interpreter $G$
    }
    \Output{Final predictions and interpreted key features of each sample and each class. Samples with high-confidence pseudo labels. Well-trained $M_1$ and $M_2$.}
    \textbf{\% Step 0: Model Pre-training} \\
    Train $M_1$ and $M_2$ on the labeled pre-training set \\
    \While{True}
    {
         \For{Each sample (or batch of samples) in $X$}
         {
               \textbf{\% Step 1: Feature Extraction} \\
                Extract features of $x_i$ and predict its local pseudo label $\hat{y}_1$ by $M_1$ \\
                Group the feature embedding into a cluster by $M_2$ using \\
                $c^{(i)} = \mathop{\arg\min}\limits_{j} ||v_i - \mu_j ||^2$. \\
                Assign the label of the $j$-th cluster $c^{(i)}$ to the unlabeled data $x_i$, i.e., the global pseudo label $\hat{y}_2$. \\
                \textbf{\% Step 2: Double Verification} \\
                \textit{\textbf{//First Verification: KAM//}} \\
                Interpret key features of the cluster by $G$ \\
                Interpret key features of the instance by $L$ \\
                Calculate a similarity score $sim$ between the interpreted global and local key features by  \\
                $ sim(int_{x}(m),int_{y}(m'), k) = 2*\frac{int^{k}_{x}(m)\cap int^{k}_{y}(m')}{\left| int^{k}_{x}(m) \right|+\left| int^{k}_{y}(m') \right|}.$
                \If{The $sim > T$}
                { \textit{\textbf{//Second Verification: ToS//}} \\
                Check the predictions of $M_1$ and $M_2$ \\
                \If{$\hat{y}_1 == \hat{y}_2$}
                {Treat them as samples with reliable pseudo-labels and use them in the next round of model training\\ }
                {Final output}
                }
         }
        Retrain $M_1$ and $M_2$ on the pseudo-labeled data \\
         }
\end{algorithm}

\subsection{Global-local Knowledge Association Mechanism (KAM)} \label{subsec:KAM}
The global-local Knowledge Association Mechanism (KAM) takes advantage of both the local and global interpreters to provide more reliable interpretations of model predictions. It uses a local (or model-agnostic) interpreter $L$ to interpret the prediction of the local model $M_1$. The interpretation results indicate which features influence the model's prediction the most. Meanwhile, it adopts a global interpreter $G$ to explain the prediction of the global model $M_2$, and the interpreted key features represent the most important features of the cluster that the current instance belongs to.

Based on the results obtained by the global interpreter $G$ and local interpreter $L$, a feature similarity calculation method (as defined in Eq.\eqref{eq:similarity}) is used to further verify the reliability of the interpretation results. Following~\cite{fan2020can}, the similarity score is defined as follows.
It measures the similarity between the locally interpreted key feature set of an instance and the globally interpreted key feature set of a cluster that the instance belongs to. 
\begin{equation}
    \label{eq:similarity}
    sim(int_{x}(m),int_{y}(m'), k) = 2*\frac{int^{k}_{x}(m)\cap int^{k}_{y}(m')}{\left| int^{k}_{x}(m) \right|+\left| int^{k}_{y}(m') \right|},
\end{equation}
where $int_x(m)$ is the interpretation result of the $x$-th sample generated by the interpreter $m$, and $k$ represents the interpreted top-$k$ key features of the $x$-th sample generated by the interpreter $m$. The $int_y(m')$ has a similar meaning where $m'$ is another interpreter. Similarly, $int_x^k(m)$ represents the number of top-$k$ interpretation results of the $x$-th sample interpreted by interpreter $m$.  

The local interpretations of an instance are considered highly reliable when the feature similarity of the local and global interpretations is higher than a given Threshold.
In our case study, we utilize the DeepLift \cite{shrikumar2017learning_deeplift} as our local interpreter and PFIE \cite{fisher2019all} as our global interpreter as an example. Other local/model-agnostic or global interpreters can also be used as the local and global interpreters in KAM.
\textcolor{black}{Note that different from existing interpreters such as \cite{han2021deepaid,liu2017contextual_coin,yang2021cade}, which only interpret anomalous samples or outliers, AnomalyAID interprets both normal and anomalous samples for that both normal and anomalous samples can potentially be used to update the model in AnomalyAID, thereby influencing the entire learning and decision-making process of anomaly detection. 
Besides, in security applications, a transparent decision-making process is crucial for system operators of security applications to adopt theoretical algorithms in practice. 
}

\subsection{Two-stage Semi-supervised Learning (ToS)} \label{subsec:ToS}
\textcolor{black}{We develop a two-stage semi-supervised learning framework named ToS to effectively assign high-confidence pseudo labels to the unlabeled samples (as illustrated in Fig. \ref{fig:mainFrame}).}

Typically, for semi-supervised learning, pseudo labels are created by a pre-trained machine learning model. In contrast with most existing semi-supervised learning methods that explore pseudo labels by only once predictions, ToS executes this process twice and the confidence of these virtual labels is improved by comparing predictions of both stages. For anomaly detection systems in security domains that only have limited labeled data, ToS repeatedly adds highly confident pseudo-labeled data for re-training the model, which significantly improves the model performance. The key idea of ToS lies in the selection of high-confidence pseudo-labeled samples and thus guarantees the subsequent process of model re-training.

As shown in Fig. \ref{fig:mainFrame}, ToS contains a local machine learning model $M_1$ and a global model $M_2$. 
During the pertaining phase, $M_1$ and $M_2$ are pre-trained on a smaller labeled set, making them capable of providing virtual labels to unlabeled data. During the re-training phase, firstly, $M_1$ extracts features of given instances and makes predictions. Secondly, $M_2$ clusters the feature embedding into several clusters according to feature distances \textcolor{black}{(see Eq.\eqref{eq:cluster}). 
\begin{equation} \label{eq:cluster}
    c^{(i)} = \mathop{\arg\min}\limits_{j} ||v_i - \mu_j ||^2 
\end{equation}
where $v_i$ denotes feature embedding of an unlabeled instance $x_i$, and $\mu_j$ represents centroids of class $j$ in labeled data. $c^{(i)}$ is the cluster that has the closest distance from $v_i$ among all cluster centroids in the labeled data.}

\textcolor{black}{
\textbf{Double Verification Mechanism.}
The goal of the double verify mechanism is to further reduce false positive and miss-detection rates and improve detection accuracy through a two-layer verification process that incorporates similar feature verification in KAM and global and local prediction checks in ToS.
For the first-layer verification, as aforementioned in Section \ref{subsec:KAM}, it compares the calculated similarity score (see Eq.\eqref{eq:similarity}), denoted as $sim$, with a predefined threshold $T$. 
We consider the interpretations to be reliable if $sim>T$. To some extent, it indicates that the sample has not been modified or attacked and could be used for the next round of re-training. 
For the second-layer verification, we consider the generated pseudo labels to be confident when the local model's prediction $\hat{y}_l$ matches the global model's prediction $\hat{y}_g$.
}
The selected reliable samples and their pseudo-labels are used to retrain both the global and clustering models. The model then waits for new network flow data to arrive for further processing.

\section{Evaluation}
This section presents the experimental setup and results. The datasets are detailed in subsection~\ref{subsec:dataset}. Next, the experiment setup is introduced in subsection~\ref{subsec:setup}. Finally, in subsection~\ref{subsec:result}, we compare the proposed KAM and ToS with 13 state-of-the-art algorithms in three representative network anomaly detection datasets and analyze their results. 

\subsection{Dataset} \label{subsec:dataset}

\begin{table*}[t!]
\footnotesize 
\centering
\caption{Descriptions of three widely-used network anomaly detection datasets used in our experiments.}
\setlength\tabcolsep{2pt}
\begin{tabular}{cccccccccc}
\toprule
\multirow{2}{*}{Dataset} & \multicolumn{2}{c}{Pre-training} & \multicolumn{2}{c}{Training} & \multicolumn{2}{c}{Validation} & \multicolumn{2}{c}{Testing} \\
\cmidrule(lr){2-3} \cmidrule(lr){4-5} \cmidrule(lr){6-7} \cmidrule(lr){8-9} 
& \#Malicious & \#Benign & \#Malicious & \#Benign & \#Malicious & \#Benign & \#Malicious & \#Benign \\
\midrule
ISCXTor2016 & 691 & 150  & 41735  & 8777 & 13146 & 2850 &  14108 & 2730\\
CIC-DoHBrw-2020  &4446 & 1350  &534045 &161499 & 173402&52650& 178049  &  53800  \\
UNSW-NB15  & 560 &  1193 & 44352  & 94518&  11088&23630&  37000 & 45332 \\
\bottomrule
\end{tabular}
\label{tab:dataset}
\end{table*}

Three widely used network anomaly detection datasets, ISCXTor2016 \cite{lashkari2017characterization}, CIC-DoHBrw-2020 \cite{montazerishatoori2020detection}, and UNSW-NB15 \cite{moustafa2015unsw}, are employed to evaluate the performance of AnomalyAID. 
ISCXTor2016 contains Tor and non-Tor network flows in various applications. Each instance has 28 features, including Tor and non-Tor flows across various applications, supporting privacy and security research. It consists of 68,191 samples, of which 56,534 are non-Tor and 11,657 are Tor samples. 
CIC-DoHBrw-2020 contains labeled traffic generated by DNS over HTTPS (DoH) protocols. It is a dataset for detecting browser-based DoH activities and studying secure communication anomalies. Each instance has 33 features. There are 933,189 instances, including 216,649 DoH and 716,540 non-DoH samples. 
The UNSW-NB15 dataset includes 100 GB of network traffic captured using tcpdump, containing nine types of attacks and 43 features generated by 12 algorithms. It is designed to support security research on modern network activities and attacks. There are 257,673 instances, including 93,000 benign and 164,673 attack samples. 

For the ISCXTor2016 dataset and UNSW-NB15 dataset, the first 1\% of the data is used as the pre-training set (i.e., small labeled set), the last 20\% as the testing set, and the next 20\% as the validation set, while the remaining data is allocated for re-training. The CSIC-DoHBrw-2020 dataset follows a similar partitioning structure, except that the first 0.5\% of the data is used as the pre-training set. Detailed data distributions of the three datasets for pre-training, training, validation, and testing are presented in Table \ref{tab:dataset}.

Note that although in this work we only evaluated AnomalyAID on network anomaly detection tasks as an example, the proposed KAM and ToS can be utilized in a wide range of semi-supervised anomaly detection applications. 

\subsection{Setup} \label{subsec:setup}
\textbf{System Settings.}
The experiments are conducted using Python (v3.8), along with the PyTorch framework (v1.11.0), CUDA (v11.3), and Scikit-learn (v1.3.2) to develop and evaluate AnomalyAID. A three-layer FCN is employed as our $M_1$, and each layer contains 256 neurons.

During training, the Adam optimizer is used with a learning rate of $1e-3$. A dropout rate of 0.3 is implemented to avoid over-fitting, and cross-entropy loss was selected as the objective function for classification tasks. 
We set the batch size to 64. DBSCAN was employed with an epsilon value of 0.5 and a minimum sample size of 1 for clustering, ensuring that all samples could potentially be included in a cluster.
The output of FCN's third fully connected layer is utilized as the input of DBSCAN. 
Additionally, early stopping was implemented with a patience of 5 epochs during both the pre-training and re-training phases so that the training process could be halted if there was no improvement in the validation loss over 5 epochs.
In PFIE, each feature is permuted 100 times. For DeepLift, the eps parameter is configured to $1e-10$. The number of key features is specified as 10, and the threshold for the similarity score is established at 0.6 from extensive experiments.

All simulations are executed on a Linux server powered by an AMD EPYC 7282 16-Core Processor (2.80 GHz) and 377 GB of RAM. To ensure consistency, all algorithms are run within the same environment, and for fairness, the average results from 10 runs are recorded for comparison.

\textbf{Benchmarks.} 
We employ a comprehensive set of benchmarks in this work to validate the performance of AnomalyAID. Firstly, we conduct a comparative analysis of the proposed KAM against eight state-of-the-art interpreters, including \textbf{Lime} \cite{ribeiro2016should_lime}, \textbf{SHAP} \cite{lundberg2017unified_shap}, \textbf{DeepLift} \cite{shrikumar2017learning_deeplift}, \textbf{COIN} \cite{liu2017contextual_coin}, 
\textbf{DiCE} \cite{mothilal2020explaining}, \textbf{xNIDS} \cite{wei2023xnids}, \textbf{CADE} \cite{yang2021cade}, \textbf{Anchors} \cite{ribeiro2018anchors}, and CETP \cite{lin2024cetp}.

\begin{itemize}
    \item{\textbf{Lime}}~\footnote{https://github.com/marcotcr/lime}: It explains the predictions of any classifier by learning an interpretable model in the local vicinity of the prediction.
    \item{\textbf{SHAP}}~\footnote{https://github.com/shap/shap}: It assigns each feature an important value using principles from cooperative game theory.
    \item{\textbf{DeepLift}}~\footnote{https://github.com/pytorch/captum}: It is an interpreter that decomposes a neural network's output prediction for a specific input by backpropagating the contributions of all neurons in the network to each feature of the input.
    \item{\textbf{COIN}}~\footnote{https://github.com/xuhongzuo/outlier-interpretation}: It transforms outlier detection into local classification tasks, identifying key features and assigning outlier scores to explain the differences between outliers and their surrounding normal instances.
    \item{\textbf{DiCE}}~\footnote{https://github.com/microsoft/DiCE}: It generates diverse counterfactual explanations by optimizing for both proximity to the original input and diversity among examples.
    \item{\textbf{xNIDS}}~\footnote{https://github.com/CactiLab/code-xNIDS}: It approximates and samples around historical inputs, and captures feature dependencies of structured data to provide high-fidelity explanations.
    \item{\textbf{CADE}}~\footnote{https://github.com/whyisyoung/CADE}: It identifies features that cause the largest changes in distance between a drifting sample and its nearest class within a low-dimensional latent space learned through contrastive learning.
    \item{\textbf{Anchors}}~\footnote{https://github.com/SeldonIO/alibi}: It provides rules that sufficiently "anchor" the prediction locally, ensuring that changes to the rest of the feature values do not affect the prediction.
    \item{\textbf{DeepSHAP}}: It combines DeepLIFT and Shapley values to compute feature attribution.
\end{itemize}

Then, we compared the introduced ToS with five state-of-the-art semi-supervised methods, including \textbf{InstantT} \cite{li2024instantT}, \textbf{ACR} \cite{wei2023towards_ACR}, \textbf{Clustering*} \cite{Fini_2023_CVPR_clustering}, \textbf{LVM} \cite{min2024leveraging_LVM},  and \textbf{M3S} \cite{sun2020multi_M3S}.

\begin{itemize}
\item{\textbf{InstantT}}~\footnote{https://github.com/tmllab/2023\_NeurIPS\_InstanT}: It assigns pseudo labels to unlabeled data by using a threshold-based approach that dynamically adjusts instance-dependent thresholds in response to label noise and prediction confidence.
    \item{\textbf{ACR}}~\footnote{https://github.com/Gank0078/ACR}: It dynamically refines pseudo-labels across various distributions by estimating the true class distribution of unlabeled data using a unified equation.
    \item{\textbf{Clustering*}}: It utilizes a multi-task framework that combines a supervised objective using ground-truth labels with a semi-supervised objective based on clustering assignments, and optimized through a single cross-entropy loss.
    \item{\textbf{LVM}}: It incorporates the dimension of local variance into pseudo-label selection.
    \item{\textbf{M3S}}: It is a multi-stage training framework that combines the DeepCluster technique with an alignment mechanism in the embedding space to enhance the training process.
    \item{\textbf{CETP}}: It assigns pseudo-labels to unlabeled data through confidence thresholds and dynamic loss weighting strategies, thereby optimizing classification performance.
\end{itemize}

\textbf{Evaluation Metrics.}
We use Accuracy (Acc.), Precision (Pre.), Recall (Rec.), F1-score (F1.), False Alarm Rate (FAR), Missed Detection Rate (MDR), and standardized partial AUC (SPAUC)
to measure different semi-supervised models' performance. Particularly, SPAUC measures the performance of a model within a specific region of the ROC curve, focusing on a segment that is most relevant under conditions of class imbalance and differential costs of misclassification. FAR reflects the amount of normal network traffic that is wrongly classified as abnormal, while MDR indicates how much anomaly network traffic that are miss-detected by the models. The higher the SPAUC, the better the performance is. The lower the FAR and MDR, the better the performance is. Here we use $\textit{SPAUC}_{\textit{FPR} \leq 0.05}$ for all experiments.

Additionally, we evaluate the performance of the proposed KAM interpreter in terms of fidelity, stability, robustness, efficiency, and AOPC, and compare it with the state-of-the-art interpreters.

\begin{itemize}
    \item{\textbf{Fidelity.}}
    To evaluate the fidelity of an interpreter, we define an indicator similar to that used in \cite{han2021deepaid}, called the Label Flipping Rate (LFR), which is the ratio of samples that change to a different class after being modified based on the interpretation results. 
    Specifically, in the context of a classification task, the interpreter identifies several key features. The values of these key features are replaced with the average values corresponding to the opposite class, and a new prediction is generated to determine whether the label has been changed. The LFR is mathematically expressed as:
    \begin{equation}
        \label{eq:fidelity}
        \textit{LFR}=\frac{\sum_{i=1}^{n} I(f'(x_i) \neq f(x_i))}{n}
    \end{equation}
    where $n$ denotes the total number of samples, $f(x_i)$ represents the original predicted label for a sample $x_i$, $f'(x_i)$ is the predicted label after modification, and $I(\cdot)$ is an indicator function that equals 1 if the label has flipped and 0 otherwise. This metric captures the percentage of samples for which the predicted label changes after altering the key features.
    \item{\textbf{Stability.}}
    The stability of interpreters refers to the consistency of interpretation results for the same samples across multiple runs. Given two interpretation outcomes $int_x(m)_1$ and $int_x(m)_2$, their similarity can be calculated using Eq.\eqref{eq:similarity}. This process is repeated to assess the similarity between the interpretation results from two runs under identical settings for each sample, and the overall stability is calculated as follows:
    \begin{equation}
        \label{eq:stability}
        Stability = \frac{1}{n} \sum_{i=1}^{n} \text{sim}(int_x(m)_1^i, int_x(m)_2^i)
    \end{equation}
    In this equation, $n$ is the total number of samples, and $\text{sim}(int_x(m)_1^i, int_x(m)_2^i)$ represents the similarity between the two interpretation outcomes for the same sample $i$. The average similarity across all samples is calculated to quantify the stability of the interpreter.
    \item{\textbf{Robustness.}}
    Robustness refers to the similarity of interpretation results for the same sample before and after adding noise, sampled from a Gaussian distribution $N(0, \sigma^2)$. Specifically, for each sample $i$, the robustness is calculated as the average similarity between the interpretation result without noise, $int_x(m)_i$, and the interpretation result after adding noise, $int_x(m + \epsilon)_i$, where $\epsilon \sim N(0, \sigma^2)$. The overall robustness is computed using the equation:
    \begin{equation}
        Robustness = \frac{1}{n} \sum_{i=1}^{n} \text{sim}(int_x(m)_i, int_x(m + \epsilon)_i, k)
    \end{equation}
    Here, $k$ represents the number of key features. In this work, the noise is set to $\sigma = 0.1$.
    \item{\textbf{Efficiency.}}
    We assess efficiency by recording the runtime required to interpret 2000 samples for each interpreter.
\end{itemize}

\subsection{Results and Analysis} \label{subsec:result}
\begin{table}[t!]
\footnotesize
  \caption{Performance (\%) of ToS on ISCXTor2016: comparisons between model performance (\%) with only the pre-training set (\ding{172}) and with both pre-training and re-training set (\ding{173}). $\triangledown$ denotes the performance difference relative to the former. The best metric performance is bolded.}
  \label{tb:PretrainPerformanceTor}
    \center
    \setlength\tabcolsep{6pt}
  \begin{tabular}{cccccccc}
    \toprule
    Setting & Acc. & Pre. & Rec. & F1. & FAR &  MDR  &  SPAUC \\
    \midrule
    \ding{172} & 80.88 & 65.68 & 67.04  & 66.29  &  12.47  &  \textbf{53.44} & 53.50 \\
    \ding{173} & \textbf{89.43} & \textbf{85.36} & \textbf{71.97} &  \textbf{76.27}  &\textbf{ 2.18} & 53.88 & \textbf{67.42}\\
    \midrule
    $\triangledown$ & +8.55 & +19.68 & +4.93  & +9.98  &  -10.29   & -0.44   & +13.92  \\
  \bottomrule
\end{tabular}
\end{table}

\begin{table}[t!]
\footnotesize
  \caption{Performance (\%) of ToS on CIC-DoHBrw-2020: comparisons between model performance (\%) with only the pre-training set (\ding{172}) and with both pre-training and re-training set (\ding{173}). $\triangledown$ denotes the performance difference relative to the former. The best metric performance is bolded.}
  \label{tb:PretrainPerformanceDoH}
    \center
    \setlength\tabcolsep{6pt}
  \begin{tabular}{cccccccc}
    \toprule
    Setting & Acc. & Pre. & Rec. & F1. & FAR &  MDR  &  SPAUC \\
    \midrule
    \ding{172} & 74.67 & 65.68 & 67.36 & 66.35 &  18.99 &  46.27 & 52.34  \\
    \ding{173} & \textbf{84.44} & \textbf{78.19} & \textbf{82.86} & \textbf{79.95}  &  \textbf{14.18} & \textbf{20.07} & \textbf{55.94} \\
    \midrule
    $\triangledown$ & +9.77 & +12.51 & +15.50  & +13.60  &  -4.81   & -26.20   & +3.60  \\
  \bottomrule
\end{tabular}
\end{table}

\begin{table}[t!]
\footnotesize
  \caption{Performance (\%) of ToS on UNSW-NB15: comparisons between model performance (\%) with only the pre-training set (\ding{172}) and with both pre-training and re-training set (\ding{173}). $\triangledown$ denotes the performance difference relative to the former. The best metric performance is bolded.}
  \label{tb:PretrainPerformanceUNSW-NB15}
    \center
    \setlength\tabcolsep{6pt}
  \begin{tabular}{cccccccc}
    \toprule
    Setting & Acc. & Pre. & Rec. & F1. & FAR &  MDR  &  SPAUC \\
    \midrule
    \ding{172} & 72.93 & 72.65 & 72.56 & 72.60  & 31.17  & \textbf{23.70}  & 51.85 \\
    \ding{173} & \textbf{78.38} & \textbf{79.38} &\textbf{78.38} & \textbf{78.38}  & \textbf{10.73}  & 30.49  & \textbf{57.02} \\
    \midrule
    $\triangledown$ & +5.45 & +6.73 & +5.82  & +5.78  & -20.44 & +6.79  & +5.17 \\
  \bottomrule
\end{tabular}
\end{table}

\begin{figure*}[t!]
    \centering
    \includegraphics[width=\textwidth]{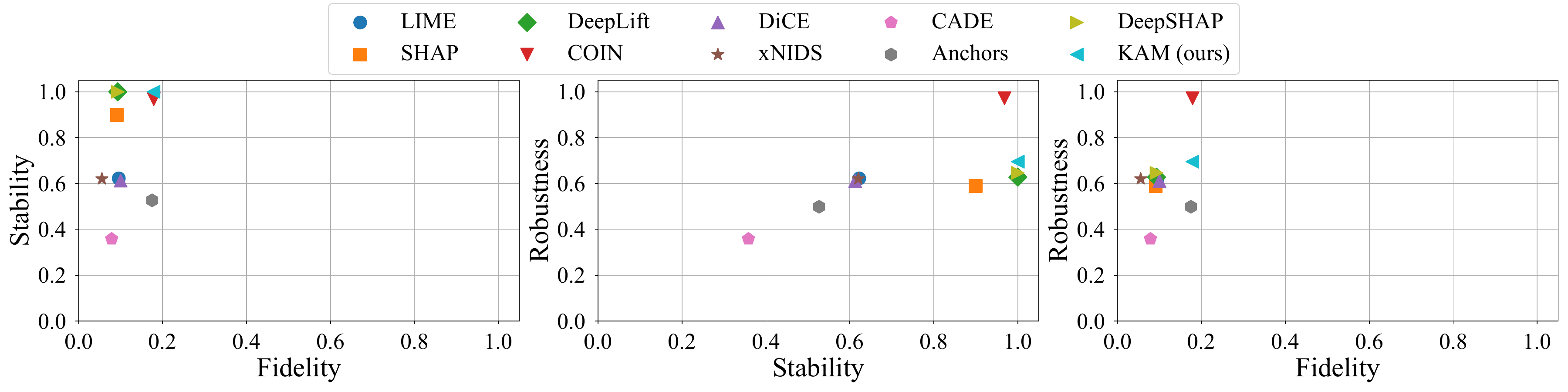}
    \caption{Fidelity, stability, and robustness evaluation of interpreters on ISCXTor2016.}
    \label{fi:interpretess_tor}
\end{figure*}

\begin{figure*}[t!]
    \centering
    \includegraphics[width=\textwidth]{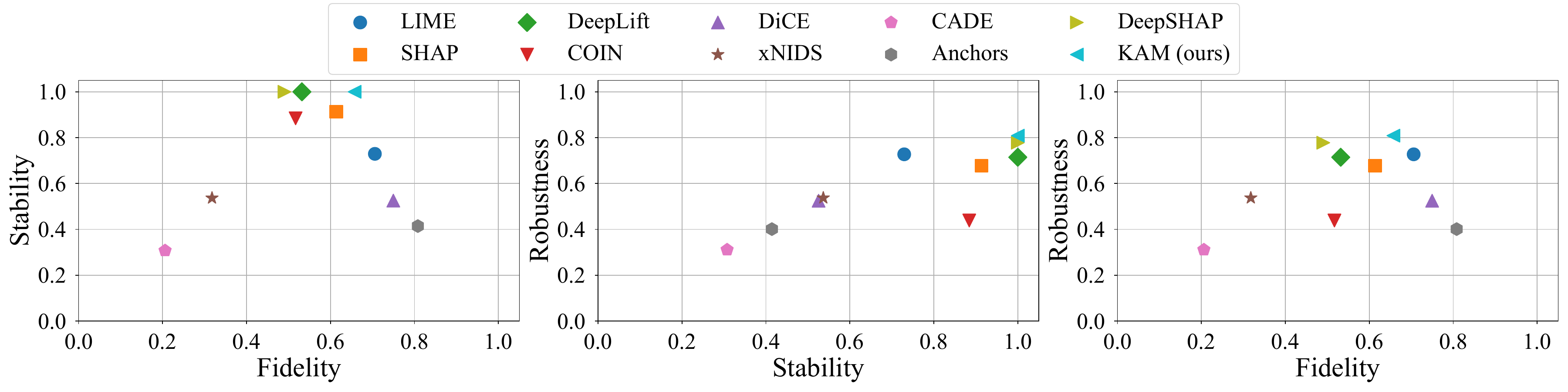}
    \caption{Fidelity, stability, and robustness evaluation of interpreters on CIC-DoHBrw-2020.}
    \label{fi:interpretess_doh}
\end{figure*}

\begin{figure*}[t!]
    \centering
    \includegraphics[width=\textwidth]{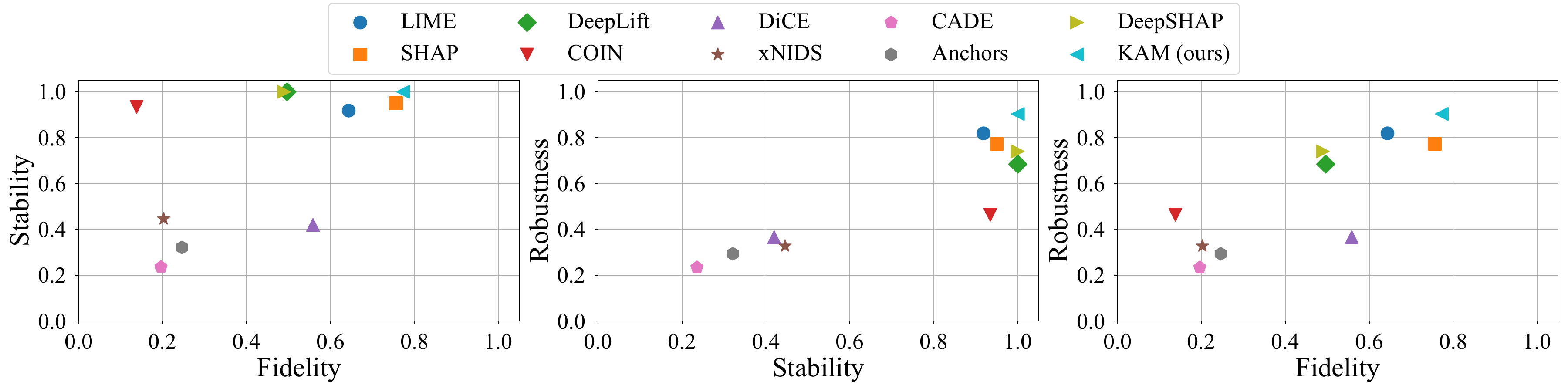}
    \caption{Fidelity, stability, and robustness evaluation of interpreters on UNSW-NB15.}
    \label{fi:interpretess_unsw}
\end{figure*}

\subsubsection{Verification of Model Pre-training.}
Table \ref{tb:PretrainPerformanceTor}, Table \ref{tb:PretrainPerformanceDoH}, and Table \ref{tb:PretrainPerformanceUNSW-NB15} present the performance of ToS when trained on the pre-training set (denoted as \ding{172}) and when trained on both the pre-training and re-training sets (denoted as \ding{173}). 
The results indicate that the re-training approach (\ding{173}) achieves significantly better performance than pre-training alone (\ding{172}) on most of the evaluation metrics in the ISCXTor2016, CIC-DoHBrw-2020, and UNSW-NB15 datasets. Specifically, it achieves 67.42\%, 55.94\%, and 57.02\% SPAUC on the ISCXTor2016, CIC-DoHBrw-2020, and UNSW-NB15 dataset, with improvements of approximately 13.92\%, 3.60\%, and 5.17\% in SPAUC compared with \ding{172}.
In particular, the performance of ToS on setting \ding{173} is 20\%, 10\%, and 14\% better in Precision, FAR, and SPAUC, on the ISCXTor2016 dataset (as presented in Table \ref{tb:PretrainPerformanceTor}). Similarly, on the CIC-DoHBrw-2020 dataset, the re-training process (\ding{173}) enhances the model's performance by about 14\%, 26\%, and 4\% in terms of F1-score, MDR, and SPAUC, compared to training only on the smaller pre-training set (\ding{172}). Additionally, as shown in Table \ref{tb:PretrainPerformanceUNSW-NB15}, the re-training process (\ding{173}) improves the model’s performance by approximately 5.45\% in accuracy, 6.73\% in precision, 5.82\% in F1-score, 5.78\% in recall, 20.44\% in FAR, and enhances the SPAUC by 5.17\%, compared to training exclusively on the pre-training set (\ding{172}) in the UNSW-NB15 dataset. Overall, ToS performs better when trained on both the pre-training and re-training sets (denoted as \ding{173}) compared to when it only trained on the pre-training set (denoted as \ding{172}), despite that the MDR of the latter (\ding{172}) is 0.44\% and 6.79\% worse than the former (\ding{173}) on the ISCXTor2016 and UNSW-NB15 datasets. 
These findings demonstrate the effectiveness of the semi-supervised learning process (i.e., ToS) in improving the performance of the proposed AnomalyAID.

\begin{figure*}[t!]
    \centering
    \includegraphics[width=\textwidth]{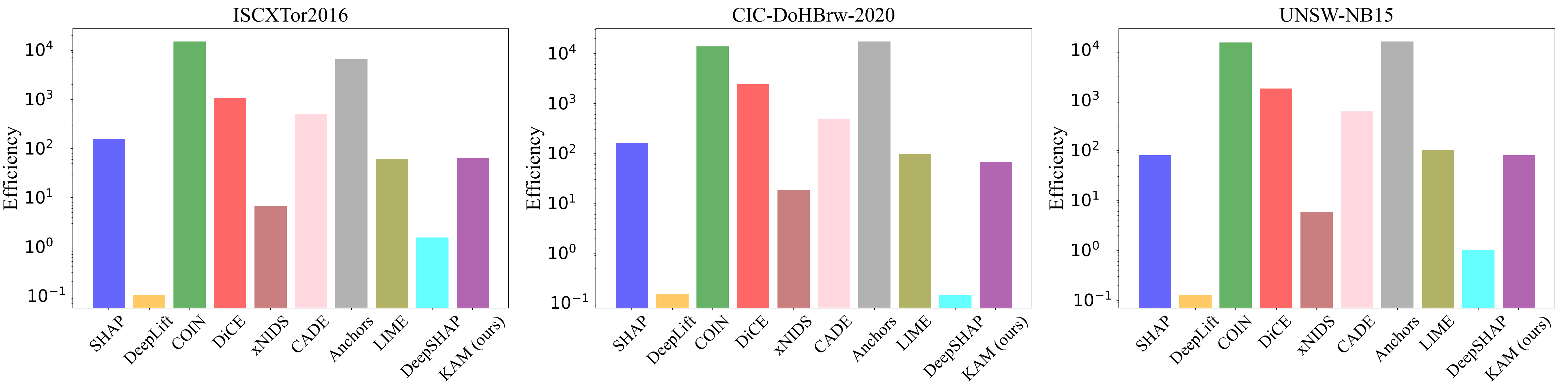}
    \caption{Efficiency evaluation of interpreters.}
    \label{fi:effiency}
    \vspace{-2mm}
\end{figure*}

\subsubsection{Comparison with the State-of-the-art Interpreters.}
Fig. \ref{fi:interpretess_tor}, Fig. \ref{fi:interpretess_doh}, and Fig. \ref{fi:interpretess_unsw} illustrate the evaluation of various interpreters across the ISCXTor2016, CIC-DoHBrw-2020, and UNSW-NB15 datasets, highlighting performance differences across three critical dimensions: fidelity, stability, and robustness. Specifically, our interpreter achieves the highest fidelity on the ISCXTor2016 and UNSW-NB15 datasets. Although some interpreters achieve higher fidelity on the CIC-DoHBrw-2020 dataset, their fidelity is lower than ours on other datasets. A higher fidelity suggests that our method effectively identifies critical features that have a substantial influence on the model's decision-making process. In terms of stability, our interpreter consistently produces similar results across multiple runs, ensuring reliable interpretations. Additionally, our method demonstrates strong robustness, maintaining accurate interpretations even when noise is introduced to the data. Overall, compared to other state-of-the-art interpreters, our model provides stable and reliable decision-making with higher credibility. Besides, Fig. \ref{fi:effiency}, in turn, presents the results of the efficiency evaluation on all three datasets. KAM demonstrates a superior balance across all these dimensions compared to other methods on all the evaluation datasets. 


As shown in Fig. \ref{fi:interpretess_tor}, we observe that in the ISCXTor2016 dataset, KAM achieves the highest fidelity compared to LIME, DeepLift, DiCE, CADE, DeepSHAP, SHAP, xNIDS, and Anchors, while the stability of KAM is higher than that of LIME, DiCE, CADE, SHAP, xNIDS, and Anchors. Additionally, KAM's robustness is higher than that of LIME, DeepLift, DiCE, CADE, DeepSHAP, SHAP, xNIDS, and Anchors. Although the robustness of COIN is higher than KAM on the ISCXTor2016 dataset, its stability is slightly worse than KAM, and its efficiency is significantly lower, being nearly 100 times slower than KAM.
Notably, the fidelity of all interpreters is relatively low on ISCXTor2016 because anomaly samples, such as Tor traffic, are more susceptible to label flipping, as modifications to their key features are more likely to alter model predictions. However, the dataset's significant class imbalance—reflected in a normal-to-anomalous ratio of approximately 4.85:1—leads to a lower overall LFR. 
In contrast, as depicted in Fig. \ref{fi:interpretess_doh} and Fig. \ref{fi:interpretess_unsw}, the CIC-DoHBrw-2020 and UNSW-NB15 dataset has a normal-to-anomalous ratio of about 3.31:1 and 0.56:1, offering a relatively more balanced scenario for evaluating fidelity.

As presented in Fig. \ref{fi:interpretess_doh}, KAM's robustness is optimal compared with all the baselines on the CIC-DoHBrw-2020 dataset. Besides, the stability of KAM outperforms LIME, DiCE, CADE, SHAP, COIN, xNIDS, and Anchors, and its fidelity is also higher compared to DeepLift, CADE, DeepSHAP, SHAP, COIN, and xNIDS. Despite that, LIME, DiCE, and ANchors obtain a higher fidelity than KAM; their stability is way worse compared to KAM. Although the stability of DeepLift is similar to that of KAM, its robustness is lower than that of KAM. KAM and DeepSHAP share a similar stability while KAM's fidelity is much better than DeepSHAP.

Fig. \ref{fi:interpretess_unsw} demonstrates the evaluation results of KAM compared with the state-of-the-art interpreters regarding fidelity, stability, and robustness on the UNSW-NB15 dataset. The results show that KAM has the best performance regarding fidelity, robustness, and stability compared to all the state-of-the-art interpreters. The stability of DeepSHAP and DeepLift is similar to KAM; however, their fidelity and robustness are much lower than KAM.

Moreover, according to Fig. \ref{fi:effiency}, we can see that KAM is more efficient than SHAP, COIN, DiCE, CADE, and Anchors on the ISCXTor2016 dataset. The efficiency of LIME is similar to that of KAM, while its stability, fidelity, and robustness are worse than KAM on ISCXTor2016. DeepLift, xNIDS, and DeepSHAP show better efficiency than KAM; however, their fidelity and robustness are much lower than KAM on ISCXTor2016. 
In addition, on the CIC-DoHBrw-2020 dataset, the efficiency of KAM is much higher than SHAP, COIN, DiCE, CADE, Anchors, and LIME, and KAM's efficiency is better than COIN, DiCE, CADE, Anchors, and LIME on the UNSW-NB15 dataset.
KAM's efficiency is not as good as that of DeepLift, xNIDS, and DeepSHAP on both CIC-DoHBrw-2020 and UNSW-NB15 datasets due to the two-branch design of KAM (i.e., with both global and local interpreters for higher reliable interpretations), but it has a higher fidelity and robustness on CIC-DoHBrw-2020 and UNSW-NB15. Even though SHAP has similar efficiency compared to KAM on the UNSW-NB15 dataset, its fidelity and robustness are way worse than KAM.  


\subsubsection{Comparison with the State-of-the-art Semi-supervised Models.} 
As shown in Table \ref{tb:comparisonLatestonTor}, Table \ref{tb:comparisonLatestonDoHBrw}, and Table \ref{tb:comparisonLatestonUNSW-NB15}, ToS outperforms other methods on the ISCXTor2016, CIC-DoHBrw-2020, and UNSW-NB15 datasets. Overall, ToS achieves SPAUC as high as 67.42\%, 55.94\%, and 57.02\% on ISCXTor2016, CIC-DoHBrw-2020, and UNSW-NB15 datasets with about 1.33\%, 1.53\%, and 3.81\% improvement, respectively. 
The accuracy, precision, recall, F1-score, and SPAUC of ToS are the best on the ISCXTor2016, CIC-DoHBrw-2020, and UNSW-NB15 datasets compared with the state-of-the-art semi-supervised models (i.e., instantT, ACR, Clustering*, LVM, M3S, and CETP). Although ToS's FAR is slightly worse than Clustering*, its accuracy, precision, recall, F1-score, MDR, and SPAUC are 6.38\%, 35.69\%, 22.23\%, 29.95\%, 45.09\%, and 17.44\% better than those of Clustering*. Similarly, instantT's MDR is 0.58\% lower than ToS, but its accuracy, precision, recall, F1-score, FAR, and SPAUC are 5.13\%, 14.54\%, 2.83\%, 6.36\%, 6.24\%, and 11.59\% worse than ToS on the ISCXTor2016 dataset, as illustrated in Table \ref{tb:comparisonLatestonTor}.

As depicted in Table \ref{tb:comparisonLatestonDoHBrw}, a similar phenomenon is observed for the CIC-DoHBrw-2020 dataset. In particular, the FAR of instantT and the MDR of ACR are better than ToS, but they do not perform as well as ToS on other evaluation metrics on the CIC-DoHBrw-2020 dataset. Similarly, ToS outperforms the state-of-the-art semi-supervised models in terms of accuracy, precision, recall, F1-score, FAR, and SPAUC. Despite the fact that instantT's FAR and ACR's MDR are lower than those of ToS, ToS significantly outperforms them in terms of other metrics on the CIC-DoHBrw-2020 dataset.

As presented in Table \ref{tb:comparisonLatestonUNSW-NB15}, the comparison results on the UNSW-NB15 dataset demonstrate that ToS archives the best SPAUC and F1-score for detecting anomalies from browser-based DoH activities compared with the state-of-the-art semi-supervised models. In particular, ToS's SPAUC is 5.58\%, 4.29\%, 4.68\%, 5.52\%, 5.40\%, and 4.46\% higher than that of instantT, ACR, Clustering*, LVM, M3S, and CETP on the UNSW-NB15 dataset. Although the accuracy, precision, and MDR of ToS are worse than instantT, its FAR and SPAUC are 36.12\% and 5.58\% better than those of instantT.

\begin{figure*}[t!]
    \centering
    \includegraphics[width=\textwidth]{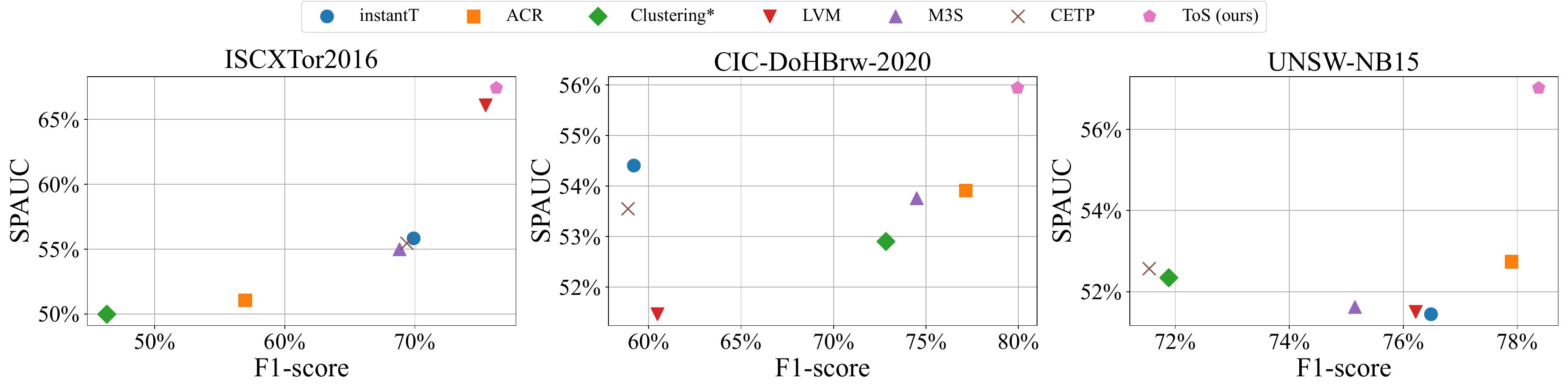}
    \caption{Performance of ToS compared with the state-of-the-art semi-supervised models: F1-score vs. SPAUC.}
    \label{fi:dot_spauc_f1}
\end{figure*}
\begin{figure*}[t!]
    \centering
    \includegraphics[width=\textwidth]{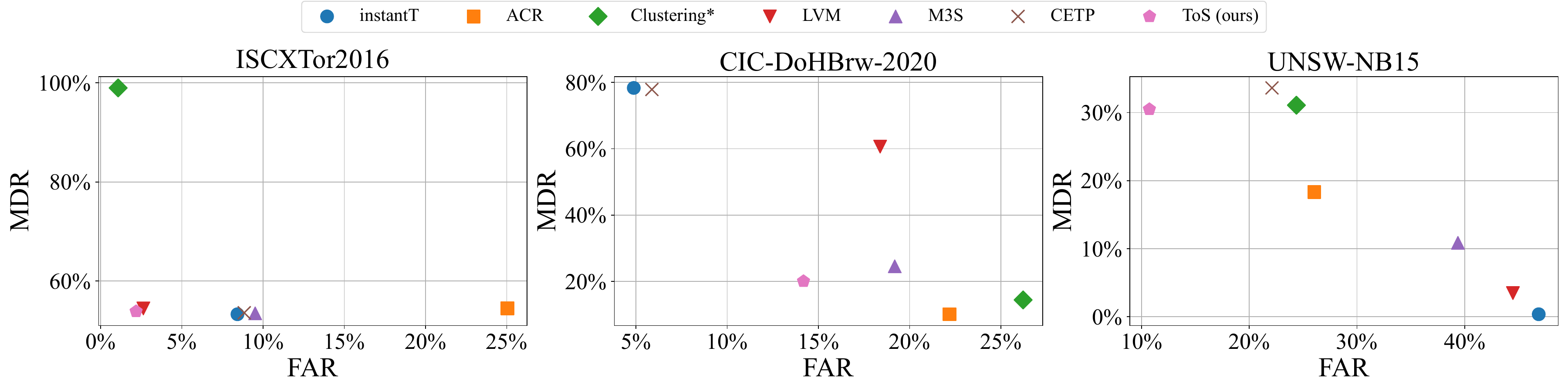}
    \caption{Performance of ToS compared with the state-of-the-art semi-supervised models: FAR vs. MDR.}
    \label{fi:dot_far_mdr}
\end{figure*}

Additionally, as illustrated in Fig. \ref{fi:dot_spauc_f1} and Fig. \ref{fi:dot_far_mdr}, ToS strikes the optimal balance between F1-score, SPAUC, FAR, and MDR. Specifically, ToS is optimal for anomaly detection on ISCXTor2016, CIC-DoHBrw-2020, and UNSW-NB15 in terms of F1-score and SPAUC compared with all the state-of-the-art approaches, as presented in Fig. \ref{fi:dot_spauc_f1}. As illustrated in Fig. \ref{fi:dot_far_mdr}, ToS outperforms the comparative state-of-the-art semi-supervised algorithms regarding FAR and MDR on the ISCXTor2016 dataset, and it has the lowest FAR on the UNSW-NB15 dataset. Although the instantT method exhibits an exceptionally low FAR on the CIC-DoHBrw-2020 dataset, its MDR is extremely high, nearing 80\%. Similarly, while ACR demonstrates superior MDR performance compared to ToS on the CIC-DoHBrw-2020 dataset, both suffer from significantly elevated FAR values.

\begin{table}[t!]
\footnotesize
  \caption{Performance (\%) of ToS compared with the state-of-the-art semi-supervised models on ISCXTor2016.}
  \label{tb:comparisonLatestonTor}
    \center
    \setlength\tabcolsep{6pt}
  \begin{tabular}{cccccccc}
    \toprule
    Model & Acc. & Pre. & Rec. & F1. & FAR &  MDR  &  SPAUC \\
     \midrule
    instantT & 84.30 & 70.82  & 69.14  & 69.91  & 8.42  & \textbf{53.30}  & 55.83  \\
    ACR & 70.18 &  56.84 &  60.23 & 56.96  & 25.04  & 54.51  & 51.05  \\
    Clustering* & 83.05 &  49.67 & 49.74  & 46.32  &  \textbf{1.07} & 98.97  & 49.98  \\
    LVM & 88.97 & 83.62  &  71.45 &  75.45 & 2.63  & 54.47  &  66.09 \\
    M3S & 83.56 &  69.18 &  68.50 &  68.83 &  9.51 &  53.48 & 54.99  \\
    CETP & 83.90  & 70.09  &  68.78  &  69.39  & 8.84  & 53.59 & 55.45  \\
    ToS (ours) & \textbf{89.43} & \textbf{85.36} & \textbf{71.97} &  \textbf{76.27}  &2.18 & 53.88 & \textbf{67.42}   \\
  \bottomrule
\end{tabular}
\end{table}

\begin{table}[t!]
\footnotesize
  \caption{Comparison with the state-of-the-art semi-supervised models' performance (\%) on CIC-DoHBrw-2020.}
  \label{tb:comparisonLatestonDoHBrw}
    \center
    \setlength\tabcolsep{6pt}
  \begin{tabular}{cccccccc}
    \toprule
    Model & Acc. & Pre. & Rec. & F1. & FAR &  MDR  &  SPAUC \\
     \midrule
    instantT & 78.07 & 68.68  & 58.40  &  59.20 & \textbf{4.89}  & 78.32  & 54.41  \\
    ACR & 80.61 &  75.62 &  83.84 & 77.15  &  22.19 &  \textbf{10.12} &  53.91 \\
    Clustering* & 76.51 & 72.03  & 79.67  & 72.84  & 26.22  &  14.43 & 52.90  \\
    LVM & 71.81 &  60.47 & 60.49 & 60.48 &  18.39 & 60.41  & 51.46  \\
    M3S & 79.56 &  72.94 &  78.12 &  74.50 &  19.19 &  24.57 & 53.76  \\
    CETP & 77.41  &  66.61 & 58.13  &  58.88  &  58.79  & 77.85  & 53.54 \\
    ToS (ours)  & \textbf{84.44} & \textbf{78.19} & \textbf{82.86} & \textbf{79.95}  &  14.18 & 20.07 & \textbf{55.94}  \\
  \bottomrule
\end{tabular}
\end{table}

\begin{table}[t!]
\footnotesize
  \caption{Comparison with the state-of-the-art semi-supervised models' performance (\%) on UNSW-NB15.}
  \label{tb:comparisonLatestonUNSW-NB15}
    \center
    \setlength\tabcolsep{6pt}
  \begin{tabular}{cccccccc}
    \toprule
    Model & Acc. & Pre. & Rec. & F1. & FAR &  MDR  &  SPAUC \\
     \midrule
    instantT & \textbf{78.74} & \textbf{85.72} & 76.39 & 76.48 & 46.85 & \textbf{0.35}  & 51.44   \\
    ACR &  78.20 &  78.02 & 77.81  & 77.89 &  26.03 & 20.64  & 52.73 \\
    Clustering* & 71.92  & 72.05 &  72.26 & 71.88  & 24.37 & 31.09  & 52.34 \\
    LVM & 78.10  & 82.78 & 76.03  &  76.21  &  44.46  & 3.47  & 51.50 \\
    M3S & 76.33  & 77.76 &  74.89  & 75.15   &  39.36  & 10.85  & 51.62 \\
    CETP & 71.54  & 72.01  & 72.13   & 71.54   &  22.11  & 33.63  & 52.56 \\
    ToS (ours)  & 78.38 & 79.38 & \textbf{78.38} & \textbf{78.38} & \textbf{10.73}  & 30.49  & \textbf{57.02}    \\
  \bottomrule
\end{tabular}
\end{table}

\section{Related Work}
In this section, we briefly discuss existing literature from two aspects.

\textbf{Interpreters.} Numerous studies have been conducted to interpret machine learning models and achieved promising results, which can be roughly categorized into two groups, i.e., local/model-agnostic \cite{ribeiro2016should,ribeiro2018anchors,lundberg2017unified_shap,guo2018_lemna,fong2019understanding} and global interpreters \cite{fisher2019all,9817459}.
The former interprets individual predictions for a given machine learning model, while the latter attempts to explain model predictions of a cluster. 
Local/model-agnostic interpreters are mostly perturbation-based, Gradient-based, and others.
Perturbation-based approaches \cite{fong2019understanding,fong2017interpretable,yang2021cade} perturb data features and check the variation of model predictions to identify the most important ones.
Gradient-based approaches \cite{shrikumar2017learning_deeplift} back-propagate gradients through the model to evaluate each feature's sensitivity.
Others treat the target machine learning model as a black-box \cite{ribeiro2016should_lime,ribeiro2018anchors}.
Techniques like LIME \cite{ribeiro2016should_lime}, LIMNA \cite{guo2018_lemna}, and SHAP \cite{lundberg2017unified_shap} attempt to utilize a linear model to approximate the decision boundary of the input, and utilize it to find the most influential features.
Global interpreters like \cite{fisher2019all} explain model predictions by randomly permuting specific feature values of a cluster and observer model predictions.
The authors of \cite{9817459} developed an interpreter for explaining clustering algorithms by formulating clustering decisions as being functionally equivalent neural networks.

Few studies aim to interpret unsupervised models such as COIN \cite{liu2017contextual_coin} and CADE \cite{yang2021cade}, which are used as baselines in our experiments. In addition to them, most of the other works directly use the interpreters proposed for supervised learning to unsupervised learning applications.

There are a few studies that examine the reliability of existing interpreters \cite{yeh2019fidelity,fan2020can}, for example, the authors of \cite{yeh2019fidelity} introduced two metrics to evaluate interpreters, i.e., sensitivity and infidelity. 
Fan et al. \cite{fan2020can} introduced three evaluation metrics to measure interpreters' stability, effectiveness, and robustness.
The work of \cite{9230374} developed criteria for evaluating and comparing different interpreters, including both general properties like accuracy and security-related ones, such as robustness, efficiency, and completeness.
Unfortunately, to our best knowledge, there is no existing work exploring how to improve the reliability of interpretation results.

\textbf{Semi-supervised Learning.} 
A recent surge of interest has focused on anomaly detection with semi-supervised learning, which attempts to assign high-confidence pseudo-labels to unlabeled data. There is a wide range of research on semi-supervised learning, which generally falls into two categories, i.e., single-stage-based and multi-stage-based approaches.
Some single-stage-based semi-supervised learning techniques improve the confidence of pseudo labels based on thresholds \cite{li2024instantT,min2024leveraging_LVM}, and they usually set high thresholds to unlabeled samples to prevent the occurrence of incorrect pseudo labels, or contrastive learning \cite{Fini_2023_CVPR_clustering,wei2023towards_ACR} that combines combine pseudo labeling and consistency regularization to encourages similar predictions between two different views of an instance to improve the robustness of the model. Unlike single-stage-based semi-supervised learning techniques, multi-stage-based semi-supervised learning approaches, such as \cite{sun2020multi_M3S,sun2020multi}, explore the most confident pseudo labels by using multiple models and improve their confidence by aligning the predictions from each stage.

\section{Conclusion}
In this paper, we propose AnomalyAID, a general framework aiming to interpret and improve semi-supervised anomaly detection performance. It incorporates two key techniques, KAM and ToS: (1) KAM provides reliable interpretations for semi-supervised systems; (2) ToS makes it possible to automatically learn from large unlabeled datasets. By applying and evaluating Adaptive NAD over two classical network anomaly detection datasets, we demonstrate that AnomalyAID can provide fast and accurate anomaly detection results as well as reliable interpretations.
As part of future work, researchers can further investigate the performance of AnomalyAID on various anomaly detection scenarios with different model combinations under extreme data distributions.

\section*{Acknowledgment}
This work was supported in part by the National Natural Science Foundation of China (Grant No. 62406215) and in part by the supported by the "Fundamental Research Funds for the Central Universities".



\bibliographystyle{elsarticle-num} 
\bibliography{main}

@article{lin2024cetp,
  title={CETP: A novel semi-supervised framework based on contrastive pre-training for imbalanced encrypted traffic classification},
  author={Lin, Xinjie and He, Longtao and Gou, Gaopeng and Yu, Jing and Guan, Zhong and Li, Xiang and Guo, Juncheng and Xiong, Gang},
  journal={Computers \& Security},
  volume={143},
  pages={103892},
  year={2024},
  publisher={Elsevier}
}

@inproceedings{moustafa2015unsw,
  title={UNSW-NB15: a comprehensive data set for network intrusion detection systems (UNSW-NB15 network data set)},
  author={Moustafa, Nour and Slay, Jill},
  booktitle={2015 military communications and information systems conference (MilCIS)},
  pages={1--6},
  year={2015},
  organization={IEEE}
}

@inproceedings{sun2020multi,
  title={Multi-stage self-supervised learning for graph convolutional networks on graphs with few labeled nodes},
  author={Sun, Ke and Lin, Zhouchen and Zhu, Zhanxing},
  booktitle={Proceedings of the AAAI conference on artificial intelligence},
  volume={34},
  number={04},
  pages={5892--5899},
  year={2020}
}

@inproceedings{shrikumar2017learning,
  title={Learning important features through propagating activation differences},
  author={Shrikumar, Avanti and Greenside, Peyton and Kundaje, Anshul},
  booktitle={International conference on machine learning},
  pages={3145--3153},
  year={2017},
  organization={PMlR}
}

@article{rumelhart1986learning,
  title={Learning internal representations by error propagation, parallel distributed processing, explorations in the microstructure of cognition, ed. de rumelhart and j. mcclelland. vol. 1. 1986},
  author={Rumelhart, David E and Hinton, Geoffrey E and Williams, Ronald J},
  journal={Biometrika},
  volume={71},
  number={599-607},
  pages={6},
  year={1986}
}

@INPROCEEDINGS{9230374,
  author={Warnecke, Alexander and Arp, Daniel and Wressnegger, Christian and Rieck, Konrad},
  booktitle={2020 IEEE European Symposium on Security and Privacy (EuroS\&P)}, 
  title={Evaluating Explanation Methods for Deep Learning in Security}, 
  year={2020},
  volume={},
  number={},
  pages={158-174},
  doi={10.1109/EuroSP48549.2020.00018}}

@article{li2024instantT,
  title={Instant: Semi-supervised learning with instance-dependent thresholds},
  author={Li, Muyang and Wu, Runze and Liu, Haoyu and Yu, Jun and Yang, Xun and Han, Bo and Liu, Tongliang},
  journal={Advances in Neural Information Processing Systems},
  volume={36},
  year={2024}
}

@inproceedings{wei2023towards_ACR,
  title={Towards realistic long-tailed semi-supervised learning: Consistency is all you need},
  author={Wei, Tong and Gan, Kai},
  booktitle={Proceedings of the IEEE/CVF Conference on Computer Vision and Pattern Recognition},
  pages={3469--3478},
  year={2023}
}

@inproceedings{min2024leveraging_LVM,
  title={Leveraging Local Variance for Pseudo-Label Selection in Semi-supervised Learning},
  author={Min, Zeping and Bai, Jinfeng and Li, Chengfei},
  booktitle={Proceedings of the AAAI Conference on Artificial Intelligence},
  volume={38},
  number={13},
  pages={14370--14378},
  year={2024}
}

@InProceedings{Fini_2023_CVPR_clustering,
    author    = {Fini, Enrico and Astolfi, Pietro and Alahari, Karteek and Alameda-Pineda, Xavier and Mairal, Julien and Nabi, Moin and Ricci, Elisa},
    title     = {Semi-Supervised Learning Made Simple With Self-Supervised Clustering},
    booktitle = {Proceedings of the IEEE/CVF Conference on Computer Vision and Pattern Recognition (CVPR)},
    month     = {June},
    year      = {2023},
    pages     = {3187-3197}
}

@ARTICLE{9817459,
  author={Kauffmann, Jacob and Esders, Malte and Ruff, Lukas and Montavon, Grégoire and Samek, Wojciech and Müller, Klaus-Robert},
  journal={IEEE Transactions on Neural Networks and Learning Systems}, 
  title={From Clustering to Cluster Explanations via Neural Networks}, 
  year={2024},
  volume={35},
  number={2},
  pages={1926-1940},
  doi={10.1109/TNNLS.2022.3185901}}

@inproceedings{mothilal2020explaining,
  title={Explaining machine learning classifiers through diverse counterfactual explanations},
  author={Mothilal, Ramaravind K and Sharma, Amit and Tan, Chenhao},
  booktitle={Proceedings of the 2020 conference on fairness, accountability, and transparency},
  pages={607--617},
  year={2020}
}

@inproceedings{wei2023xnids,
  title={$\{$xNIDS$\}$: Explaining Deep Learning-based Network Intrusion Detection Systems for Active Intrusion Responses},
  author={Wei, Feng and Li, Hongda and Zhao, Ziming and Hu, Hongxin},
  booktitle={32nd USENIX Security Symposium (USENIX Security 23)},
  pages={4337--4354},
  year={2023}
}

@inproceedings{yang2021cade,
  title={$\{$CADE$\}$: Detecting and explaining concept drift samples for security applications},
  author={Yang, Limin and Guo, Wenbo and Hao, Qingying and Ciptadi, Arridhana and Ahmadzadeh, Ali and Xing, Xinyu and Wang, Gang},
  booktitle={30th USENIX Security Symposium (USENIX Security 21)},
  pages={2327--2344},
  year={2021}
}

@inproceedings{liu2017contextual_coin,
  title={Contextual outlier interpretation},
  author={Liu, Ninghao and Shin, Donghwa and Hu, Xia},
  booktitle={Proceedings of the 27th International Joint Conference on Artificial Intelligence},
  pages={2461--2467},
  year={2018}
}

@article{yeh2019fidelity,
  title={On the (in) fidelity and sensitivity of explanations},
  author={Yeh, Chih-Kuan and Hsieh, Cheng-Yu and Suggala, Arun and Inouye, David I and Ravikumar, Pradeep K},
  journal={Advances in neural information processing systems},
  volume={32},
  year={2019}
}

@article{fan2020can,
  title={Can we trust your explanations? Sanity checks for interpreters in Android malware analysis},
  author={Fan, Ming and Wei, Wenying and Xie, Xiaofei and Liu, Yang and Guan, Xiaohong and Liu, Ting},
  journal={IEEE Transactions on Information Forensics and Security},
  volume={16},
  pages={838--853},
  year={2020},
  publisher={IEEE}
}

@inproceedings{ribeiro2016should,
  title={" Why should i trust you?" Explaining the predictions of any classifier},
  author={Ribeiro, Marco Tulio and Singh, Sameer and Guestrin, Carlos},
  booktitle={Proceedings of the 22nd ACM SIGKDD international conference on knowledge discovery and data mining},
  pages={1135--1144},
  year={2016}
}

@article{guidotti2018survey,
  title={A survey of methods for explaining black box models},
  author={Guidotti, Riccardo and Monreale, Anna and Ruggieri, Salvatore and Turini, Franco and Giannotti, Fosca and Pedreschi, Dino},
  journal={ACM computing surveys (CSUR)},
  volume={51},
  number={5},
  pages={1--42},
  year={2018},
  publisher={ACM New York, NY, USA}
}

@INPROCEEDINGS{9155278,
  author={Tang, Ruming and Yang, Zheng and Li, Zeyan and Meng, Weibin and Wang, Haixin and Li, Qi and Sun, Yongqian and Pei, Dan and Wei, Tao and Xu, Yanfei and Liu, Yan},
  booktitle={IEEE INFOCOM 2020 - IEEE Conference on Computer Communications}, 
  title={ZeroWall: Detecting Zero-Day Web Attacks through Encoder-Decoder Recurrent Neural Networks}, 
  year={2020},
  pages={2479-2488}}

@inproceedings {259729,
author = {Benjamin Bowman and Craig Laprade and Yuede Ji and H. Howie Huang},
title = {Detecting Lateral Movement in Enterprise Computer Networks with Unsupervised Graph {AI}},
booktitle = {23rd International Symposium on Research in Attacks, Intrusions and Defenses (RAID 2020)},
year = {2020},
isbn = {978-1-939133-18-2},
address = {San Sebastian},
pages = {257--268},
publisher = {USENIX Association},
month = oct
}

@inproceedings{mirsky2018kitsune,
  title={Kitsune: An Ensemble of Autoencoders for Online Network Intrusion Detection},
  author={Mirsky, Yisroel and Doitshman, Tomer and Elovici, Yuval and Shabtai, Asaf},
  booktitle={25th Annual Network and Distributed System Security Symposium, NDSS 2018},
  year={2018},
  organization={The Internet Society}
}

@article{fisher2019all,
  title={All Models are Wrong, but Many are Useful: Learning a Variable's Importance by Studying an Entire Class of Prediction Models Simultaneously.},
  author={Fisher, Aaron and Rudin, Cynthia and Dominici, Francesca},
  journal={J. Mach. Learn. Res.},
  volume={20},
  number={177},
  pages={1--81},
  year={2019}
}

@inproceedings{lashkari2017characterization,
  title={Characterization of tor traffic using time based features},
  author={Lashkari, Arash Habibi and Gil, Gerard Draper and Mamun, Mohammad Saiful Islam and Ghorbani, Ali A},
  booktitle={International Conference on Information Systems Security and Privacy},
  volume={2},
  pages={253--262},
  year={2017},
  organization={SciTePress}
}

@inproceedings{sun2020multi_M3S,
  title={Multi-stage self-supervised learning for graph convolutional networks on graphs with few labeled nodes},
  author={Sun, Ke and Lin, Zhouchen and Zhu, Zhanxing},
  booktitle={Proceedings of the AAAI conference on artificial intelligence},
  volume={34},
  number={04},
  pages={5892--5899},
  year={2020}
}

@inproceedings{ribeiro2016should_lime,
  title={"Why should i trust you?" Explaining the predictions of any classifier},
  author={Ribeiro, Marco Tulio and Singh, Sameer and Guestrin, Carlos},
  booktitle={Proceedings of the 22nd ACM SIGKDD international conference on knowledge discovery and data mining},
  pages={1135--1144},
  year={2016}
}

@article{lundberg2017unified_shap,
  title={A unified approach to interpreting model predictions},
  author={Lundberg, Scott M and Lee, Su-In},
  journal={Advances in neural information processing systems},
  volume={30},
  number={1},
  year={2017}
}

@inproceedings{guo2018_lemna,
  title={Lemna: Explaining deep learning based security applications},
  author={Guo, Wenbo and Mu, Dongliang and Xu, Jun and Su, Purui and Wang, Gang and Xing, Xinyu},
  booktitle={proceedings of the 2018 ACM SIGSAC conference on computer and communications security},
  pages={364--379},
  year={2018}
}

@inproceedings{fong2017interpretable,
  title={Interpretable explanations of black boxes by meaningful perturbation},
  author={Fong, Ruth C and Vedaldi, Andrea},
  booktitle={Proceedings of the IEEE international conference on computer vision},
  pages={3429--3437},
  year={2017}
}

@inproceedings{fong2019understanding,
  title={Understanding deep networks via extremal perturbations and smooth masks},
  author={Fong, Ruth and Patrick, Mandela and Vedaldi, Andrea},
  booktitle={Proceedings of the IEEE/CVF international conference on computer vision},
  pages={2950--2958},
  year={2019}
}

@inproceedings{shrikumar2017learning_deeplift,
  title={Learning important features through propagating activation differences},
  author={Shrikumar, Avanti and Greenside, Peyton and Kundaje, Anshul},
  booktitle={International conference on machine learning},
  pages={3145--3153},
  year={2017},
  organization={PMLR}
}

@inproceedings{ribeiro2018anchors,
  title={Anchors: High-precision model-agnostic explanations},
  author={Ribeiro, Marco Tulio and Singh, Sameer and Guestrin, Carlos},
  booktitle={Proceedings of the AAAI conference on artificial intelligence},
  volume={32},
  number={1},
  year={2018}
}

@inproceedings{han2021deepaid,
  title={DeepAID: Interpreting and Improving Deep Learning-based Anomaly Detection in Security Applications},
  author={Han, Dongqi and Wang, Zhiliang and Chen, Wenqi and Zhong, Ying and Wang, Su and Zhang, Han and Yang, Jiahai and Shi, Xingang and Yin, Xia},
  booktitle={Proceedings of the 2021 ACM SIGSAC Conference on Computer and Communications Security},
  pages={3197--3217},
  year={2021}
}

@inproceedings{ester1996density,
  title={A density-based algorithm for discovering clusters in large spatial databases with noise.},
  author={Ester, Martin and Kriegel, Hans-Peter and Sander, J{\"o}rg and Xu, Xiaowei and others},
  booktitle={kdd},
  volume={96},
  number={34},
  pages={226--231},
  year={1996}
}

@INPROCEEDINGS{montazerishatoori2020detection,
  author={MontazeriShatoori, Mohammadreza and Davidson, Logan and Kaur, Gurdip and Habibi Lashkari, Arash},
  booktitle={2020 IEEE Intl Conf on Dependable, Autonomic and Secure Computing, Intl Conf on Pervasive Intelligence and Computing, Intl Conf on Cloud and Big Data Computing, Intl Conf on Cyber Science and Technology Congress (DASC/PiCom/CBDCom/CyberSciTech)}, 
  title={Detection of DoH Tunnels using Time-series Classification of Encrypted Traffic}, 
  year={2020},
  pages={63-70}}





\end{document}